\documentclass{article}

\PassOptionsToPackage{numbers,sort&compress}{natbib}
\usepackage[preprint]{neurips_2026}
\workshoptitle{2nd Workshop on Grounded and Faithful Vision Language Models for Real-World Deployment}
\usepackage[T1]{fontenc}
\usepackage[utf8]{inputenc}
\usepackage{times}
\usepackage{microtype}
\usepackage{graphicx}
\usepackage{booktabs}
\usepackage{multirow}
\usepackage{makecell}
\usepackage{adjustbox}
\usepackage{amsmath,amssymb,mathtools}
\usepackage{xcolor}
\usepackage{colortbl}
\usepackage{xspace}
\usepackage{enumitem}
\usepackage{caption}
\usepackage{float}
\usepackage{placeins}
\usepackage{url}
\usepackage[hidelinks]{hyperref}
\usepackage{cleveref}

\definecolor{methodcolor}{HTML}{006B9E}
\definecolor{softblue}{HTML}{EEF5FA}
\definecolor{softgray}{HTML}{F2F2F2}
\newcommand{\method}{\textcolor{methodcolor}{\textbf{\textsc{Before The Flip}}}\xspace}
\newcommand{\prs}{\operatorname{PRS}}
\newcommand{\lfaith}{\mathcal{L}_{\mathrm{faith}}}
\newcommand{\recovery}{\Delta^{F}}

\newcommand{\yes}{\textsc{yes}}
\newcommand{\no}{\textsc{no}}

\renewcommand{\paragraph}[1]{\smallskip\noindent\textbf{#1.}}
\setlist[itemize]{leftmargin=1.35em,itemsep=1pt,topsep=2pt,parsep=0pt}
\title{\method{}: Measuring Hidden Score Shifts 
\\
In Quantized Vision Language Models Before The Answer Changes for Visual Question Answering}

\author{%
Sourajit Saha \quad Shubhashis Roy Dipta \quad Shaswati Saha \quad Nobin Sarwar \quad Yuxuan Jiang\\[3pt]
University of Maryland, Baltimore County\\
{\tt\small \{ssaha2, sroydip1, ssaha3, sms2, yuxuanj1\}@umbc.edu}\\
{\textcolor{blue}{\normalsize\url{https://sourajitcs.github.io/beforetheflip/}}}\\[2pt]
}

\begin{document}

\maketitle

\begin{center}
    \includegraphics[width=\linewidth]{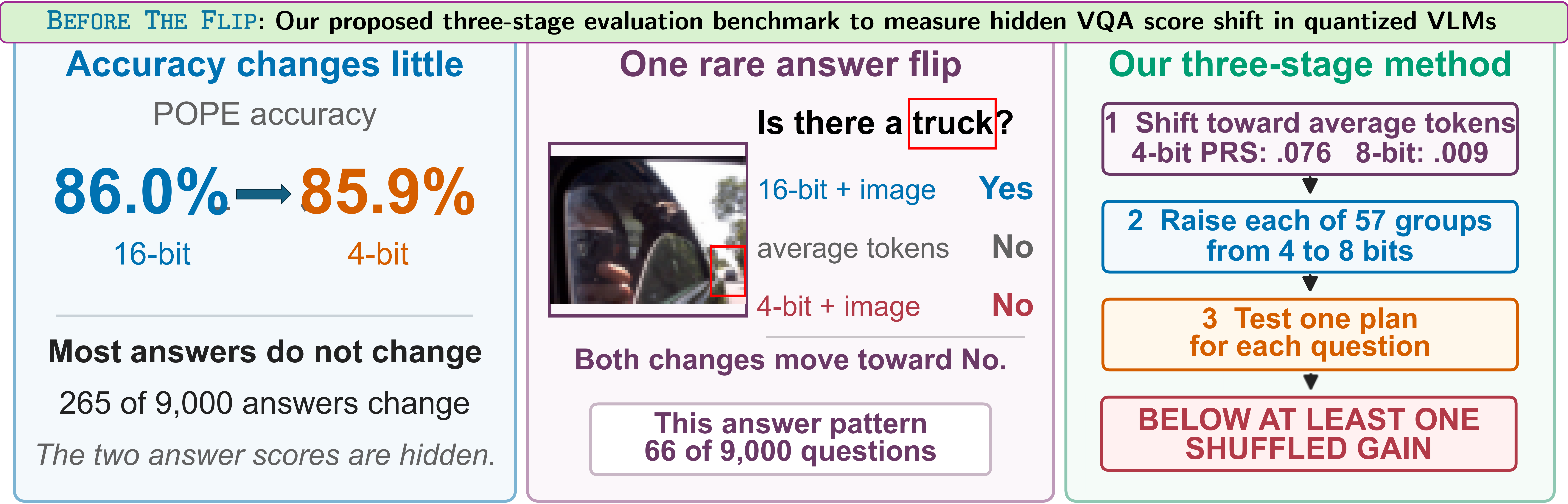}
    \captionof{figure}{
    \textbf{Visual Question Answering (VQA) Accuracy alone can hide the effects of model compression.}
    Although 4-bit quantization barely changes VQA accuracy and flips only $265$ of $9{,}000$ answers, it can still shift the underlying yes/no scores toward the output produced with average image tokens.
    \method{} measures this hidden shift, identifies which weight groups cause it, and tests whether selecting higher precision groups for each question provides a reliable benefit.
    The example answer flip pattern occurs in $66$ of $9{,}000$ questions, while per question selection fails to outperform both shuffled controls on $1{,}024$ separate calibration questions.
    }
    \label{fig:teaser}
\end{center}
\vspace{1em}
\begin{abstract}
Quantization makes \emph{vision language models} (VLMs) cheaper to store and run by using fewer bits to represent their weights. While unchanged answers on \emph{visual question answering} (VQA) after compression are an expected behavior, they can still hide changes in the underlying scores (log probabilities). For example, a model may still answer ``yes'' after compression, even as the score gap between ``yes'' and ``no'' shrinks. We introduce \method{} to measure these hidden changes. Our method compares the score change caused by compression with the change caused by replacing the image's internal representations, or image tokens, with one fixed average token. We then increase the precision of one weight group at a time to identify where extra bits help, and test whether choosing different groups for each question offers benefits beyond shuffled controls. Among $8{,}277$ LLaVA questions where image token replacement measurably affects the scores, 4-bit compression shifts the yes/no score gap farther toward the replacement output than 8-bit compression. Qwen shows the same pattern, but with a smaller difference. Yet only $265$ of $9{,}000$ LLaVA answers change at 4 bits. In a separate study of $1{,}024$ calibration questions, choosing weight groups separately for each question does not outperform both shuffled controls at any tested storage budget. These findings show that compression can alter the scores behind unchanged answers, but do not establish a reliable benefit from adjusting precision for each question.
\end{abstract}

\section{Introduction}
\label{sec:introduction}

The goal of quantizing a vision language model (VLM) is to make it cheaper to store and run without changing its answers on \emph{visual question answering} (VQA) among other tasks. Quantization does this by storing each weight with fewer bits. Consider the question, ``Is there a truck in the image?'' The model gives one score to \yes{} and one score to \no{}, otherwise known as \emph{logits} converted to \emph{log probabilities}. An accuracy test sees only the answer with the larger score. It never measures changes in the gap between the two scores of \yes{} and \no{}. This gap can move even when the answer stays identical. Similar hidden changes appear in other settings, where the final output still looks correct even though an internal preference or a dropped constraint has already changed~\citep{hossain2026right, mazumder2026agentcollabbench, sarwar2026multimodal}. Current VLM quantization studies predominantly report benchmark scores~\citep{wang2024qvlm, li2025mbq, zhang2026maba, yu2025mquant}. Current grounding tests study full precision models~\citep{li2023pope, rohrbach2018chair, leng2024vcd, sarwar2025filterrag}, leaving quantized VLMs at deployment risk for high stake AI applications across domains~\cite{saha2018lightning, kamran2019optic, saha2018total, kamran2020comprehensive, ravin2022mitigating, saha2022mypath, saha2022pairwise, sarwar2025fedmentor, saha2018efficient, joshi2024towards, saha2025side, saha2025improving, sarwar2025fedmentalcare, saha2026erase, lia-roy-dipta-2026-cross, saha2023seebel, saha2026zero}. We therefore lack principled evaluation for score changes in the model used at deployment after compression. To study this gap, we present \method{}, a three stage method for VQAs in quantized VLMs.

In Stage-1, we introduce the Prior Reversion Score (PRS) to measure how compression shifts the scores behind an answer. The first uses the input image and 16 bit weights. The second keeps the 16 bit weights but replaces all image tokens with the same fixed average token. The third uses the input image and compressed weights. PRS compares the score shift caused by compression with the shift caused by average token replacement. A positive PRS means that compression moves the scores for \yes{} and \no{} toward the average token output. The term ``prior reversion'' describes only this measured direction and does not establish the existence of a hidden language prior.

In Stage-2, we identify which parts of the model benefit from greater precision. We divide the model weights into $57$ groups and initially store every group at 4 bits. We then raise one group at a time to 8 bits and run the questions again. This reveals where additional precision reduces the measured loss.

In Stage-3, we test whether the most useful groups differ reliably across questions. Selecting groups after observing all results may appear effective simply because of measurement noise~\citep{roy-dipta-ferraro-2025-may}. We therefore compare the actual gain with two shuffled controls. One preserves the values for each group, while the other preserves the values for each question.

Our results reveal one clear pattern and one key limitation. For LLaVA, the average PRS increases from $0.0094$ at 8 bits to $0.0760$ at 4 bits, showing that stronger compression causes a larger score shift. Qwen2.5 VL shows the same pattern, but less strongly. Despite these score shifts, only $265$ of $9{,}000$ LLaVA answers change at 4 bits. Giving more precision to earlier decoder blocks reduces the loss more on average, although different parts within some blocks have opposite effects. At every tested storage budget, selecting groups separately for each question performs worse than at least one shuffled control. On separate calibration questions, all three selection methods choose the same precision setting for every question. These results reveal score changes hidden behind mostly unchanged answers, but they do not support a reliable system that adjusts precision for each question.

\noindent\textbf{Our contributions and findings are:}
\begin{itemize}
  \item \textbf{A metric for hidden score shifts.} We introduce the Prior Reversion Score (PRS) to measure whether compression moves the \yes{} and \no{} scores toward the image replacement output, even when the final answer remains unchanged after model compression via quantization.

  \item \textbf{A detailed map of precision sensitivity.} By raising one weight group at a time from 4 to 8 bits, we measure the effects of $24$ image encoder blocks, one connector, and $32$ decoder blocks.

  \item \textbf{A controlled test of per question precision selection.} We compare the observed gain with two shuffled controls when each question can use $5\%$, $10\%$, or $15\%$ more weight storage than W4. The observed gain remains below at least one shuffled control at every budget.

  \item \textbf{Evaluation across models, tasks, and settings.} We test two VLMs, five image replacement methods, open ended captioning, four additional tasks, fixed precision plans, learned selection methods, and visual contrastive decoding.
\end{itemize}
\section{Related Work}

\paragraph{Quantizing language models}
Post training serves as the continuing improvement method on LLMs' reasoning ability~\citep{jiang2026bridgingreasoningtrajectoriesonpolicy,jiang2026scribestructuredmidlevelsupervision}, where quantization reduces the number of bits used to store a trained model without training it again for the target task. GPTQ and AWQ decide how to quantize weights using quantization error or activation size~\citep{frantar2023gptq, lin2024awq}. NF4 provides 4 bit values designed for normally distributed weights~\citep{dettmers2023qlora}. LLM.int8() keeps important outlier calculations at higher precision, while SmoothQuant moves difficult activation values into the weights~\citep{dettmers2022llmint8, xiao2023smoothquant}. ZeroQuant and OmniQuant use hardware aware weight groups or learned calibration rules~\citep{yao2022zeroquant, shao2024omniquant}. SpQR and SqueezeLLM handle unusual weights separately or represent them with nonuniform values~\citep{dettmers2024spqr, kim2024squeezellm}. All these methods aim to reduce cost while preserving the original model's performance~\citep{jiang-etal-2026-drp}.

\paragraph{Quantizing vision language models}
Quantizing VLMs is more difficult because image and text features can have very different scales. Q-VLM, MBQ, MABA, and MQuant identify the model parts that are most affected by quantization~\citep{wang2024qvlm, li2025mbq, zhang2026maba, yu2025mquant}. Related analyses in language models show that the units carrying the largest measured signal are not always the ones that matter functionally~\citep{jiang2026cornerstones,yang2026modularizedreinforcementlearningllms,jiang-ferraro-2026-beyond}. VVSQ, MASQuant, and LUQ assign different levels of precision to different model parts~\citep{kim2026vvsq, hu2026masquant, bhatnagar2025luq}. QAQ and RouteLLM study related decisions based on the input query~\citep{li2025qaq, ong2025routellm, nazi2026triage, dipta2026pa3}. These studies motivate our test of whether the best precision setting depends on the question.

\paragraph{Testing whether an answer follows the image}
VQA v2 pairs similar questions with different images, while VQA-CP changes common question and answer patterns between training and testing~\citep{goyal2017vqav2, agrawal2018vqacp}. HINT tests whether a model focuses on image regions that people consider important~\citep{selvaraju2019hint}. SugarCrepe and HallusionBench use controlled examples to reveal shortcuts and hallucinations~\citep{hsieh2023sugarcrepe, guan2024hallusionbench}. POPE tests questions about objects, while CHAIR measures objects mentioned in captions but absent from the image~\citep{li2023pope, rohrbach2018chair, roy-dipta-etal-2026-vc}. VCD and M3ID compare model outputs produced with the original image and a modified image~\citep{leng2024vcd, favero2024m3id, nazi2026omni}. OPERA and HALC modify the decoding process to reduce object hallucinations~\citep{huang2024opera, chen2024halc}. DoLa compares outputs from different model depths~\citep{chuang2024dola}. Recent work also studies when VLMs ignore images and whether attention reveals how they use visual information~\citep{zhou2026visualignorance, zhang2026visualinsensitivity, song2026visualattention, sayeedi2026many}. Our work instead measures how low bit weights move the two answer scores under a controlled image replacement.

\paragraph{Choosing when to use a stronger system}
Selective prediction allows a model to reject uncertain cases~\citep{geifman2019selectivenet}. Confidence calibration tests whether a model's confidence matches its probability of being correct~\citep{guo2017calibration, hossain2026uat}. Learning to defer sends difficult cases to another decision maker~\citep{madras2018defer, narasimhan2022posthocdefer}. Our setting asks a related question: should a particular input use a more expensive precision setting? We require the gain from choosing precision separately for each question to exceed two shuffled controls. One control shuffles values across questions within each weight group. The other shuffles values across weight groups within each question.
\FloatBarrier
\section{\method{}: Our Proposed Three Stage Evaluation}
\label{sec:method}

\begin{figure}[H]
  \centering
  \includegraphics[width=\linewidth]{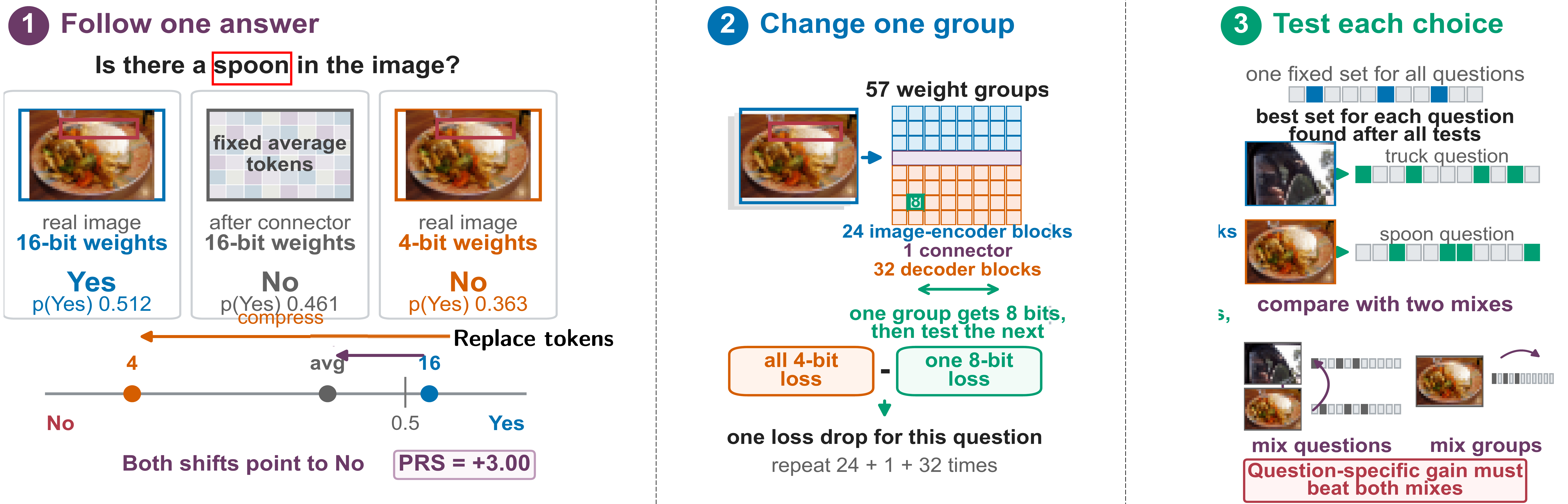}
  \caption{\textbf{Overview of \method{}.} Stage~1 measures whether compression moves the answer scores toward the output produced by average image tokens. Stage~2 raises each of $57$ weight groups from 4 to 8 bits, one at a time, and records the reduction in loss. Stage~3 tests whether selecting groups for each question outperforms one fixed group set and two shuffled controls, and determines whether those gains are reliable.}
  \label{fig:diagnostic}
\end{figure}

\subsection{Stage~1: Measure the Hidden Score Shift}

\paragraph{Three runs of the same question}
Each POPE question has two possible answers, \yes{} and \no{}. For question $i$, $z_{16,i}^{I}$ contains the two scores produced by the 16 bit model using image $I$. We subtract their average from each score. This removes a shared offset and leaves only the score gap that determines the answer: $c(z)=z-\frac{1}{2}\mathbf{1}\mathbf{1}^{\top}z$. The image replacement run keeps the 16 bit weights but replaces the image tokens, producing scores $z_{16,i}^{\emptyset}$. The compressed run uses the real image and produces scores $z_{\pi,i}^{I}$, where $\pi$ gives the bit width of each weight group. The resulting score changes:
\begin{equation}
  d_i=c(z_{16,i}^{\emptyset})-c(z_{16,i}^{I}),\qquad
  q_i^{\pi}=c(z_{\pi,i}^{I})-c(z_{16,i}^{I}).
  \label{eq:directions}
\end{equation}
Here, $d_i$ is the change caused by replacing the image tokens, while $q_i^{\pi}$ is the change caused by compression. Both changes are measured against the original 16 bit output using the input image.

\paragraph{Prior Reversion Score}
We introduce PRS to measure how much the compression change follows the image replacement change:
\begin{equation}
  \prs_i(\pi)=\frac{\langle q_i^{\pi},d_i\rangle}{\lVert d_i\rVert_2^2+\epsilon}.
  \label{eq:prs}
\end{equation}
A positive PRS means that compression moves the two answer scores toward the image replacement output. A score of zero means no movement in that direction. $\epsilon$ prevents division by zero. PRS can be unstable when replacing image tokens causes little change. We therefore keep questions with $\lVert d_i\rVert_2>\tau_d$, where $\tau_d=0.302$ is the tenth percentile measured on separate examples. We also report three measures. PRS${_+=\max(\prs,0)}$ keeps only movement toward the replacement output. The response ratio compares how strongly the compressed and 16 bit models respond to image replacement. The flip match rate counts changed answers that match the replacement answer.

\paragraph{Replacing the image tokens}
The image encoder converts an image into features, and the connector converts these features into image tokens for the decoder. We replace every token after the connector with the same fixed average token computed from separate examples. The number and positions of the tokens, the question, and the prompt remain unchanged. We repeat the analysis using an average from the same image, zero tokens, tokens from another image, and a gray input image.

\subsection{Stage~2: Change One Weight Group}

We divide the model into ${G=57}$ groups: $24$ image encoder blocks, one connector, and $32$ decoder blocks. The starting model uses 4 bit weights. For each test, we raise only group $g$ to its corresponding 8 bit weights.

Our loss contains two terms: the added cross entropy error on the correct answer and the positive part of PRS. We scale them to similar ranges:
\begin{equation}
  {\lfaith}_i(\pi)=
  \frac{[\operatorname{CE}_i(\pi)-\operatorname{CE}_i(16)]_+}{s_{\mathrm{CE}}}
  +\frac{[\prs_i(\pi)]_+}{s_{\mathrm{PRS}}}.
  \label{eq:lfaith}
\end{equation}
Here, ${[x]_+=\max(x,0)}$ (lower value is better). We define improvement from changing group $g$ as
\begin{equation}
  \recovery_{i,g}={\lfaith}_i(\mathrm{W4})-
  {\lfaith}_i(\mathrm{W4}[g{\rightarrow}\mathrm{W8}]).
  \label{eq:recovery}
\end{equation}
A positive value means that raising group $g$ to 8 bits lowers the loss for question $i$. This measures only the effect of changing that group and does not show where visual grounding is stored.

\subsection{Stage~3: Test a Choice for Each Question}

A fixed plan selects one group set using the average loss drop across questions. A hindsight plan selects a different set for each question after seeing every result. This plan cannot be deployed and may appear effective because it can select gains caused by measurement noise.

We test this possibility using two shuffled controls. Shuffle A mixes questions within each group, preserving the values for that group while breaking question identity. Shuffle B mixes groups within each question, preserving the difficulty of that question while breaking group identity. The real hindsight gain must exceed both shuffled gains to support per question selection. Finally, we run every selected group set through the model because the effects of changing multiple groups may interact.
\section{Experiments}
\label{sec:experiments}

Our experiments address three questions: (1) whether 4 bit quantization shifts the yes/no score gap farther toward the average token output than 8 bit quantization; (2) which weight groups most reduce the measured loss; and (3) whether useful precision groups can be selected reliably for individual questions. We further evaluate whether the observed patterns generalize across models, tasks, image replacements, and decoding methods.

\subsection{Experimental Setup}

\paragraph{Models and quantization}
We use LLaVA-1.5-7B as our primary model~\citep{liu2024llava15} and Qwen2.5-VL-7B-Instruct to evaluate whether the main bit width trend transfers to another model~\citep{bai2025qwen25vl}. Across all evaluated batches, we experiment with 4 bit and 8 bit quantization.

\paragraph{Data and model groups}
POPE-COCO contains $3{,}000$ random, $3{,}000$ popular object, and $3{,}000$ adversarial yes/no questions~\citep{li2023pope}. Our primary intervention replaces every image token with an average token estimated from a separate calibration set. LLaVA retains all $576$ image token positions. For Qwen, we use a fixed resolution of $448\times448$ with $256$ image tokens instead of its native variable resolution. The group analysis uses $1{,}024$ balanced questions constructed from COCO train2014~\citep{lin2014coco}. We partition the images into disjoint subsets for precision profile construction, loss limit calibration, and selector evaluation. The group map and shuffle analyses use all $1{,}024$ questions. We additionally divide $13$ decoder blocks into separate attention and feed forward groups.

\paragraph{Evaluation tasks}
We evaluate answer accuracy on all $9{,}000$ POPE questions and caption hallucination on $500$ CHAIR examples~\citep{rohrbach2018chair}. Additional evaluations cover MME perception, MMBench, ARC, and HellaSwag~\citep{fu2025mme, liu2024mmbench, zellers2019hellaswag, clark2018think}. For the text only tasks, we provide the same neutral image to every example. We also evaluate visual contrastive decoding (VCD), which adjusts answer scores by comparing outputs from real and corrupted images~\citep{leng2024vcd}.

\paragraph{Statistical analysis}
We compute confidence intervals using $10{,}000$ image level bootstrap samples and compare paired accuracy using McNemar's test. The flip match shuffle preserves both the POPE subset and answer label. We report mean PRS without an image level confidence interval.

\paragraph{Analysis subsets}
The primary PRS analysis excludes questions with a negligible response to image replacement, whereas the POPE precision plan results include all $9{,}000$ questions. For W4, mean PRS$_+$ is $0.241$ on the complete set and $0.127$ after filtering.
\FloatBarrier
\subsection{Results and Findings}

\subsubsection{Quantization Shifts Scores Before Answers Change}

\begin{table}[H]
  \centering
  \caption{
    \textbf{Compression shifts the answer scores, and the measured shift depends on the image replacement.}
    Panel (a) shows a larger PRS under W4 than W8 despite similar final performance.
    Panel (b) compares W4 across five image replacements and shows that the direction and size of PRS depend on the chosen replacement.
    PRS uses $8{,}277$ filtered questions, while final answer measures use all $9{,}000$ questions.
  }
  \label{tab:main_diagnostic}
  \vspace{5pt}
  \begin{minipage}[t]{0.56\linewidth}
    \vspace{0pt}
    \centering
    \small \textbf{(a) Score movement by precision}\par\vspace{3pt}
    \setlength{\tabcolsep}{2.2pt}
    \resizebox{\linewidth}{!}{
      \begin{tabular}{@{}lrrrrrrr@{}}
        \toprule
        Weights
        & F1 $\uparrow$
        & Acc. $\uparrow$
        & \makecell{Mean\\PRS}
        & \makecell{PRS\\${>}0$}
        & \makecell{Response\\ratio}
        & \makecell{Flip match\\real/shuffled}
        & $p$ \\
        \midrule
        8-bit
        & 84.811 & 86.067 & .0094 & 63.7\%
        & 1.008 & .455/.456 & .974 \\
        \rowcolor{softblue}
        4-bit
        & 84.722 & 85.922 & .0760 & 72.7\%
        & 1.058 & .426/.427 & .722 \\
        \bottomrule
      \end{tabular}
    }
  \end{minipage}
  \hfill
  \begin{minipage}[t]{0.43\linewidth}
    \vspace{0pt}
    \centering
    \small \textbf{(b) Sensitivity to image replacement}\par\vspace{3pt}
    \setlength{\tabcolsep}{2.5pt}
    \resizebox{\linewidth}{!}{
      \begin{tabular}{@{}lrrr@{}}
        \toprule
        Image replacement
        & \makecell{Yes\\(\%)}
        & F1
        & \makecell{W4 PRS across\\POPE splits} \\
        \midrule
        Dataset average token
        & 0.6 & 1.4 & .061 to .091 \\
        Image average token
        & 2.1 & 6.2 & .066 to .091 \\
        All zero tokens
        & 100.0 & 66.7 & .049 to .070 \\
        Another image's tokens
        & 27.8 & 64.1 & .043 to .043 \\
        Uniform gray image
        & 0.0 & 0.0 & $-.042$ to .004 \\
        \bottomrule
      \end{tabular}
    }
  \end{minipage}
\end{table}

\paragraph{W4 produces a larger score shift than W8}
Table~\ref{tab:main_diagnostic} shows that across the $8{,}277$ questions with a measurable response to image replacement, mean PRS increases from $0.0094$ under W8 to $0.0760$ under W4.
The fraction of questions with positive PRS similarly increases from $63.7\%$ to $72.7\%$.
This ordering holds across all three POPE subsets and across the tested filter thresholds.
Because an image level confidence interval is unavailable for mean PRS, we treat this comparison as a descriptive pattern.

Final answers remain considerably more stable.
W4 changes $265$ of the $9{,}000$ answers.
Of these changes, $42.6\%$ match the answer produced by average token replacement, compared with $42.7\%$ under the shuffled control ($p=.722$).
The median response ratio is $1.058$, indicating that the higher PRS is not caused by a weaker response to image replacement.
The primary effect of compression is therefore a shift in the answer scores and not in widespread answer changes.

\paragraph{PRS depends on the image replacement}
Table~\ref{tab:main_diagnostic} compares five image interventions.
All four token replacements produce positive W4 PRS in every POPE subset, whereas a gray image passed through the image encoder does not.
Their outputs also differ substantially.
Dataset average tokens produce \yes{} for only $0.6\%$ of questions, zero tokens produce \yes{} for every question, and tokens from another image produce \yes{} for $27.8\%$ with an F1 of $64.1$.
PRS therefore measures movement toward a specified replacement output and should not be interpreted as evidence of a general language prior.
Appendix~\ref{sec:app-construct} reports results for every subset and cutoff.

\FloatBarrier
\subsubsection{Higher Precision Helps Earlier Decoder Blocks on Average}

\begin{figure}[H]
  \centering
  \includegraphics[width=0.9\linewidth]{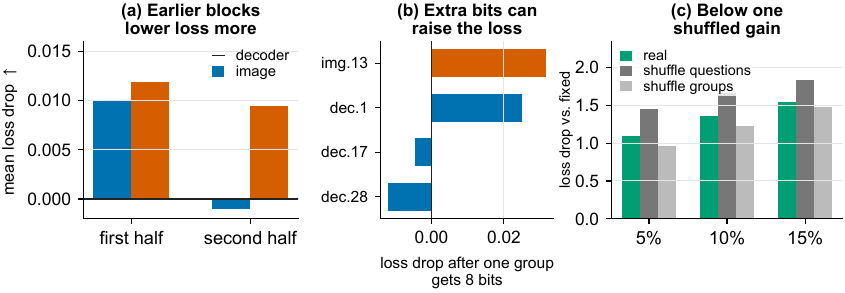}
  \caption{
    \textbf{Weight group sensitivity and shuffle controls.}
    (a) Raising earlier decoder blocks to 8 bits reduces the loss more on average.
    (b) Individual group changes can either reduce or increase the loss.
    (c) At every storage budget, the per question gain fails to exceed both shuffled controls.
  }
  \label{fig:structure_nulls}
\end{figure}

\paragraph{Loss reduction decreases with decoder depth}
Across the $32$ decoder blocks, depth is negatively correlated with average loss reduction ($\rho=-0.633$, $p=0.0001$; Figure~\ref{fig:structure_nulls}a).
The first half of the decoder has a mean loss reduction of $0.00994$, compared with $-0.00098$ for the second half.
The $24$ image encoder blocks show no measurable depth trend ($\rho=-0.133$, $p=.535$).

This trend does not hold for every block.
Thirteen decoder blocks increase the measured loss when raised to W8.
The effect can also change when a block is divided into smaller components.
For example, decoder block 17 changes from $-0.0045$ as a complete block to $+0.0042$ for attention alone.
The mean difference between a complete block and the sum of its components is $0.00203$.
We therefore evaluate every combined precision experiments directly through the model.
These interventions measure the effect of individual weight groups but do not identify where visual grounding is stored.
Appendix~\ref{sec:app-groups} provides the complete group map.

\subsubsection{Per Question Selection Does Not Beat Both Shuffled Controls}

\paragraph{Similar gains appear after shuffling}
Figure~\ref{fig:structure_nulls}c compares the observed and shuffled gains on $1{,}024$ calibration questions.
The observed gains are $1.0966$, $1.3535$, and $1.5458$ at the $5\%$, $10\%$, and $15\%$ storage budgets.
At $10\%$, the observed gain has a $95\%$ interval of $[1.2142,1.5066]$, while the two shuffled gains are $1.6616$ and $1.2309$.
It lies between them.
The same ordering occurs at $5\%$ and $15\%$, so per question selection fails to outperform both shuffled controls at every tested budget.

\paragraph{Question specific variation is not repeatable}
Group rankings computed from two halves of the questions have a correlation of only $r=.307$.
Average differences among groups explain $0.15\%$ of the measured variance, and average differences among questions explain $6.33\%$.
The remaining $93.52\%$ combines question by group variation with measurement noise, which cannot be separated using one observation per pair.
Appendix~\ref{sec:app-routing} reports the complete controls.

\subsection{Further Analysis}

\subsubsection{Fixed and Learned Precision Plans}

We next test whether learned selectors can choose precision profiles for individual questions.
Table~\ref{tab:policy_summary} summarizes the resulting plans.

\vspace{-5pt}
\begin{table}[H]
  \centering
  \caption{
    \textbf{The fixed $10\%$ plan lowers PRS$_+$ and measured loss from W4.}
    The $95\%$ interval for its PRS$_+$ difference from the image encoder rule includes zero.
    PRS$_+$ and loss use all $9{,}000$ questions. Total stored GiB includes every selector candidate.
  }
  \label{tab:policy_summary}
  \vspace{3pt}
  \small \textbf{Fixed plans and the selector choice before VCD}\vspace{2pt}
  \setlength{\tabcolsep}{5.2pt}
  \begin{tabular}{@{}lrrrrr@{}}
    \toprule
    Setting & Acc. & PRS$_+$ & Measured loss & \makecell{Bits/weight for\\one question} & \makecell{Total stored\\GiB} \\
    \midrule
    All weights 4-bit & 85.92 & .241 & .514 & 4.00 & 3.17 \\
    All weights 8-bit & 86.07 & .032 & .076 & 8.00 & 6.33 \\
    Image encoder 8-bit & 85.62 & .167 & .295 & 4.18 & 3.31 \\
    \rowcolor{softblue}
    Fixed 10\% average token plan & 85.76 & .176 & .307 & 4.39 & 3.48 \\
    Selector choice, before VCD & 85.79 & .181 & .328 & 4.38 & 3.79 \\
    \bottomrule
  \end{tabular}
\end{table}
\vspace{-5pt}

\paragraph{All selectors collapse to the same choice}
We evaluate linear regression, a small neural network, and a fixed distance between two pairs of yes/no scores.
Each selector predicts the loss of four precision profiles and adds an error margin estimated from held out data.
No upper prediction satisfies the W8 based loss limit of $\epsilon_f=0.324$.
All three selectors therefore choose K3\_P2 and invoke VCD for every held out calibration question.
Before VCD, K3\_P2 has a loss of $0.4075$, compared with $0.3557$ for the best fixed profile, K3\_P3.
The difference is $0.0518$, with a $95\%$ interval of $[-0.0065,0.1129]$.

\paragraph{Prediction error obscures differences among profiles}
Mean absolute prediction error ranges from $0.429$ to $0.642$ across the candidate profiles.
This error is comparable to the W4 to W8 loss range of $0.550$ to $0.070$ on the same held out examples.
Relaxing the loss limit allows more questions to pass but does not improve prediction accuracy.

\paragraph{The fixed plan improves over W4 but not the image encoder rule}
Across all POPE questions, the fixed $10\%$ plan reduces PRS$_+$ by $0.0650$ relative to W4, with a $95\%$ interval of $[-0.0868,-0.0424]$.
Adversarial accuracy changes by only $-0.10$ points, with an interval of $[-0.70,+0.50]$.

Relative to keeping the image encoder at W8, the fixed plan has a PRS$_+$ difference of $+0.0085$, with an interval of $[-0.0079,+0.0264]$.
Relative to keeping the first eight decoder blocks at W8, it reduces PRS$_+$ by $0.0247$, with an interval of $[-0.0476,-0.0008]$.
Uniform W4 uses $3.17$ GiB, while storing every selector candidate uses $3.79$ GiB.
All sizes are calculated from parameter counts.
Appendices~\ref{sec:app-routing} and~\ref{sec:app-profiles} report all precision plans.

\paragraph{Uniform W4 satisfies the general task criterion}
Appendix~\ref{sec:app-profiles} evaluates precision plans using MME, MMBench, ARC, and HellaSwag.
A task violates the preservation criterion only when its score falls by more than two points and two standard errors.
Uniform W4 produces no such violation and therefore requires no additional high precision groups under this criterion.

\FloatBarrier
\subsubsection{Tasks, Models, and Decoding Settings}

\begin{figure}[!t]
  \centering
  \includegraphics[width=0.9\linewidth]{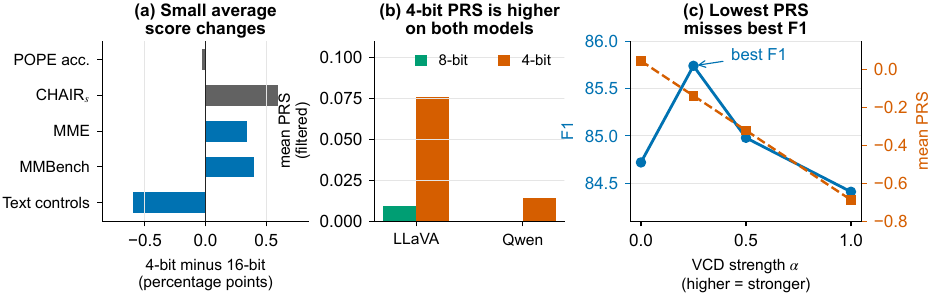}
  \caption{
    \textbf{Results across tasks, models, and decoding settings.}
    (a) W4 changes five task averages by less than one point relative to 16 bit.
    (b) W4 produces higher PRS than W8 on both LLaVA and Qwen, although the Qwen gap is smaller.
    (c) The VCD setting with the lowest mean PRS does not produce the highest F1.
  }
  \label{fig:scope}
\end{figure}

\paragraph{Aggregate task performance changes little}
Figure~\ref{fig:scope}a summarizes five task averages.
From 16 bit to W4, POPE F1 changes by $+0.0358$ points and accuracy by $-0.0333$ points.
CHAIR$_s$ increases from $24.0\%$ to $24.6\%$, with a $95\%$ interval of $[-2.80,+4.00]$ points and $p=.773$.
MME, MMBench, and the pooled text controls each change by less than one point, although ARC Easy decreases by $3.20$ points.
These results show no consistent overall degradation, but they do not establish deployment safety.
Object recall remains between $61.9\%$ and $63.9\%$, while average caption length remains between $49.3$ and $49.6$ words.
Appendix~\ref{sec:app-behavior} reports the complete task results.

\begin{figure}[!t]
  \centering
  \includegraphics[width=0.9\linewidth]{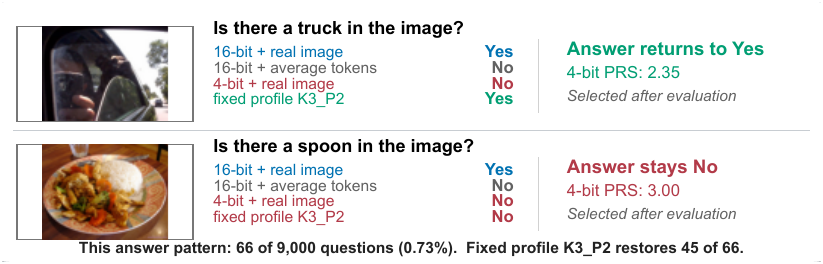}
  \caption{
    \textbf{Two examples of hidden answer changes.}
    In both cases, 16 bit predicts \yes{}, while average tokens and W4 predict \no{}.
    The fixed K3\_P2 profile restores the truck answer but not the spoon answer.
    These examples were selected after evaluating all $9{,}000$ questions.
  }
  \label{fig:qualitative}
\end{figure}

\paragraph{Qwen reproduces the bit ordering with a smaller gap}
On fixed resolution Qwen2.5-VL, mean PRS increases from $0.00017$ under W8 to $0.01397$ under W4 (Figure~\ref{fig:scope}b).
F1 decreases from $88.938$ to $88.624$, and the W4 response ratio is $0.9985$.
The flip match rate is $0.5766$, compared with $0.5372$ after shuffling ($p=.198$).
Thus, Qwen reproduces the higher PRS under W4, but the difference between W4 and W8 is much smaller than for LLaVA.

\paragraph{Optimizing PRS does not optimize VQA performance}
Without VCD, W4 has a mean PRS of $0.0458$, F1 of $84.72$.
F1 reaches $85.74$ at VCD strength $0.25$, while mean PRS continues decreasing through strength $1.0$.
At strength $1.0$, mean PRS reaches $-0.688$, but F1 falls to $84.41$, PRS$_+$ rises to $1.243$, and accuracy decreases by $2.73$ points.
Mean PRS and PRS$_+$ therefore favor different settings, and neither alone provides a complete objective for selecting VCD strength.
Because the settings were compared on evaluation data, these results do not define a selection rule.
The learned selector invokes VCD for every question and is therefore equivalent to always applying VCD.

\paragraph{The qualitative examples are selected after evaluation}
Figure~\ref{fig:qualitative} shows two of the $66$ questions, or $0.73\%$ of POPE, for which 16 bit predicts \yes{} while both average token replacement and W4 predict \no{}.
The fixed K3\_P2 profile restores $45$ of these $66$ answers.
Because the examples use one fixed profile, they illustrate the score shift but do not demonstrate successful per question selection.
\section{Conclusion}

We introduced \method{} to quantify score changes hidden by stable VQA accuracy after quantization.
On LLaVA, W4 moves the \yes{}--\no{} score margin farther toward the average token output than W8, although only $265$ of $9{,}000$ answers change.
Qwen reproduces this ordering with a smaller effect.
Raising individual groups to 8 bits benefits earlier decoder blocks on average, but per question gains fail to exceed both shuffled controls.
The selectors converge to one profile and invoke VCD for every question, while the fixed $10\%$ plan does not outperform the image encoder rule.
Thus, quantization can change internal scores before answers flip, but PRS depends on the chosen replacement and does not yet support adaptive precision.
\section{Limitations}
\label{sec:limitations}

Our experiments use low precision weights with 16 bit operations and report without latency, memory, or energy measurements.
PRS is limited to binary answers and a replacement favoring \no{}, while mean PRS lacks image level confidence intervals.
Future work should examine additional VLMs.

\clearpage
{
    \bibliographystyle{unsrtnat}
    \bibliography{references,extra}
}

\clearpage
\appendix
\section*{Appendix}
\section{Detailed Results and Definitions}
\label{sec:appendix}
\setlength{\textfloatsep}{5pt plus 1pt minus 1pt}
\setlength{\intextsep}{5pt plus 1pt minus 1pt}
\setlength{\floatsep}{5pt plus 1pt minus 1pt}
\renewcommand{\arraystretch}{0.80}

This appendix reports every completed test. W4 and W8 store model weights with 4 and 8 bits. The 16-bit model is the reference. An image-replacement run changes the image tokens but keeps the question, token count, and token positions fixed. The Prior Reversion Score (PRS) compares the movement caused by lower precision with the movement caused by image replacement. Positive PRS points toward the replacement output. Zero means no movement in that direction. A value of one reaches the replacement output along that direction. PRS$_+$ keeps only positive movement. The \emph{faithfulness loss} adds PRS$_+$ to any increase in the model's penalty on the correct answer. Lower loss is better. A separate \emph{calibration} set fixes thresholds and candidate plans before testing. A precision ``profile'' is one fixed plan for assigning weight bits. The vision encoder turns pixels into image tokens. The connector maps those tokens into the language model. The decoder produces the answer. CI means confidence interval.

The POPE results use two populations. The filtered population removes cases where image replacement changes the 16-bit scores very little. PRS is unstable in those cases. The all-question population keeps all 9,000 POPE questions when comparing precision plans. Filtered PRS must not be compared with all-question PRS$_+$ or faithfulness loss. The group and selection studies use a separate pool of 1,024 calibration questions or held-out parts of that pool. All experiments were conducted using four NVIDIA L40S GPUs.

\subsection{How the Score Depends on the Image-Replacement Test?}
\label{sec:app-construct}

\paragraph{Filtering changes the score size but not the bit ordering}
The predeclared cutoff is the tenth percentile of image-effect size. It keeps $91.97\%$ of questions. On this filtered population, average PRS is $0.07596$ for W4 and $0.00940$ for W8 (Table~\ref{tab:app-populations}). On all questions, the means are $0.04582$ and $0.00081$. W4 PRS$_+$ is $0.12701$ after filtering and $0.24101$ on all questions. A nearly zero image effect can make the division in PRS very large. The value $0.07596$ describes the directional result. The value $0.24101$ applies only to all-question policy comparisons.

\begin{table}[!ht]
  \centering
  \caption{\textbf{The stability filter changes the score size.} The first two rows remove weak image-effect cases with cutoff $\tau_d=0.302$. The last two rows keep every question. Bit widths should be compared only within the same pair of rows.}
  \label{tab:app-populations}
  \vspace{3pt}
  \setlength{\tabcolsep}{4pt}
  \begin{tabular}{llrrrr}
    \toprule
    Questions used & Weight bits & Avg. PRS & PRS$_+$ & Positive (\%) & Kept (\%) \\
    \midrule
    $\lVert d_i\rVert>\tau_d$ & W8 & 0.00940 & 0.01842 & 63.73 & 91.97 \\
    $\lVert d_i\rVert>\tau_d$ & W4 & \textbf{0.07596} & 0.12701 & 72.66 & 91.97 \\
    \midrule
    All questions & W8 & 0.00081 & 0.03247 & 62.31 & 100.00 \\
    All questions & W4 & \textbf{0.04582} & 0.24101 & 70.67 & 100.00 \\
    \bottomrule
  \end{tabular}
\end{table}

\paragraph{Nearby cutoffs keep the same ordering}
We test cutoffs at the fifth, tenth, and twentieth percentiles. W4 PRS remains positive and above W8 for every POPE question type (Table~\ref{tab:app-tau}). A stricter cutoff lowers the kept share from $94$--$96\%$ to $81$--$87\%$. W4 changes from $0.062$ to $0.059$ for random negatives. It changes from $0.077$ to $0.082$ for popular-object negatives. It changes from $0.064$ to $0.072$ for adversarial negatives. The cutoff changes the numerical value but does not create the ordering.

\begin{table}[!ht]
  \centering
  \caption{\textbf{W4 remains more positive at all three stability cutoffs.} ``Kept'' is the percentage of questions retained. Each PRS value uses only the retained questions.}
  \label{tab:app-tau}
  \vspace{3pt}
  \setlength{\tabcolsep}{5pt}
  \begin{tabular}{lcrrrrrr}
    \toprule
    Cutoff $\tau_d$ & Bits & Random & Kept & Popular & Kept & Adversarial & Kept \\
    \midrule
    5th pct. ($0.201$)  & W8 & .009 & 96 & .011 & 95 & .010 & 94 \\
                        & W4 & .062 & 96 & .077 & 95 & .064 & 94 \\
    10th pct. ($0.302$) & W8 & .009 & 94 & .010 & 91 & .009 & 91 \\
                        & W4 & .061 & 94 & .091 & 91 & .076 & 91 \\
    20th pct. ($0.581$) & W8 & .009 & 87 & .010 & 82 & .008 & 81 \\
                        & W4 & .059 & 87 & .082 & 82 & .072 & 81 \\
    \bottomrule
  \end{tabular}
\end{table}

\paragraph{The result depends on where image information is changed}
Four tests replace tokens after the connector has mapped vision output into language-model input. All four give positive W4 PRS on the three question types (Table~\ref{tab:app-replacements}). A separate test sends a gray image through the full vision encoder. It gives $0.004$, $-0.042$, and $-0.039$. The result therefore applies to replacement after the connector. It does not cover every way to change image information. Tokens from another image are the only alternative that produces both yes and no answers often. Their PRS is $0.043$ for all three question types. Wrong-image evidence therefore behaves differently from fixed-token replacement.

\begin{table}[!ht]
  \centering
  \caption{\textbf{Four token replacements agree in sign, while a gray input image does not.} Yes-rate and F1 describe the 16-bit model after replacement. The final columns report filtered W4 PRS.}
  \vspace{3pt}
  \label{tab:app-replacements}
  \setlength{\tabcolsep}{4pt}
  \begin{tabular}{lrrrrr}
    \toprule
    How image information is replaced & Yes (\%) & F1 & W4 random & W4 popular & W4 adversarial \\
    \midrule
    Calibration mean token & 0.6 & 1.4 & .061 & .091 & .076 \\
    Per-image mean token   & 2.1 & 6.2 & .066 & .091 & .081 \\
    Zero tokens            & 100.0 & 66.7 & .049 & .070 & .068 \\
    Shuffled-image tokens  & 27.8 & 64.1 & .043 & .043 & .043 \\
    Uniform gray image     & 0.0 & 0.0 & .004 & $-.042$ & $-.039$ \\
    \bottomrule
  \end{tabular}
\end{table}

\paragraph{The main replacement run does not reveal a language prior}
Replacing every image token with the calibration-set average makes the model answer yes on only $0.6\%$ of POPE. Its F1 is $1.4$. Per-image averages and the gray image also produce almost only one answer. Zero tokens produce the opposite extreme and answer yes every time. Positive PRS means movement toward the specified replacement output. It does not show movement toward an internal co-occurrence belief. ``Prior reversion'' is the name of the measurement. It is not an identified mental mechanism.

\paragraph{Changed final answers match the shuffled expectation}
W4 changes 265 of 9,000 yes/no answers. Among those changes, $0.4264$ match the 16-bit replacement answer. The average is $0.4269$ after shuffling within groups that share the true label and question type ($p=0.7225$). W8 changes only 22 answers. Its matching rate is also unresolved against the shuffle control ($0.4545$ versus $0.4563$, $p=0.9740$). The median ratio of lower-precision to 16-bit response under image replacement is $1.0578$ for W4 and $1.0078$ for W8. The answer scores show a bit ordering. Final-answer reversions and weaker image response do not.

\subsection{Which Model Parts Respond to Extra Precision}
\label{sec:app-groups}

\paragraph{Vision blocks and early decoder blocks lower loss on average}
We divide the model into 57 replaceable groups. These are 24 vision blocks, one connector, and 32 language-decoder blocks as shown in Table~\ref{tab:app-groups}. ``Recovery'' $\Delta^F$ is the drop in faithfulness loss when one group changes from W4 to W8. Positive recovery is better. Vision blocks give more recovery per stored byte. Each vision block has about 13M parameters, while a decoder block has 202M. Decoder block 1 ranks third in absolute recovery ($\Delta^F=0.0249$). Decoder blocks 13, 10, 6, and 8 also lower loss. A Spearman rank correlation measures whether the ordering changes with depth. Decoder recovery falls with depth, with $\rho=-0.633$ ($p=0.0001$). The first-half mean is $0.00994$, compared with $-0.00098$ in the second half. Vision depth has no resolved trend ($\rho=-0.133$, $p=0.5354$). Its half means are $0.01191$ and $0.00946$. Decoder block 28 has recovery $-0.0119$. Extra precision can therefore worsen the combined score.

\begin{table}[!ht]
  \centering
  \caption{\textbf{Changing one group from W4 to W8 can reduce or increase loss.} ``Loss drop'' is $\Delta^F$, and positive is better. Accuracy changes are percentage points. The table includes representative large and negative effects. No single ordering fits every block.}
  \label{tab:app-groups}
  \vspace{3pt}
  \setlength{\tabcolsep}{4pt}
  \begin{tabular}{llrrllrr}
    \toprule
    Group & Model part & Loss drop & Acc. change & Group & Model part & Loss drop & Acc. change \\
    \midrule
    vis.13 & vision  & .0316    &  .20   & dec.13 & decoder & .0223    &  .10   \\
    vis.06 & vision  & .0292    &  .29   & dec.10 & decoder & .0209    & $-.29$ \\
    dec.01 & decoder & .0249    & $-.49$ & dec.06 & decoder & .0197    &  .00   \\
    vis.08 & vision  & .0239    &  .68   & dec.08 & decoder & .0193    &  .10   \\
    vis.14 & vision  & .0189    &  .10   & dec.20 & decoder & $-.0043$ & $-.10$ \\
    dec.14 & decoder & $-.0061$ &  .00   & dec.28 & decoder & $-.0119$ &  .59   \\
    \bottomrule
  \end{tabular}
\end{table}

\paragraph{The loss-lowering subpart changes across decoder blocks}
We split 13 decoder blocks into two tested subparts. Attention mixes information across tokens. The feed-forward network (FFN) transforms each token separately. Block 1's gain comes almost entirely from the FFN ($-0.0001$ attention, $0.0275$ FFN). Block 6's gain comes from attention ($0.0217$ attention, $-0.0039$ FFN). Both subparts lower loss in block 13. Their separate gains still sum to $0.0052$ less than the whole-block gain. The mean absolute mismatch is $0.00203$. Separate effects therefore do not add exactly.

A whole-block change lowers loss in ten of the tested decoder blocks. In five of them, one subpart supplies at least $90\%$ of the whole-block loss drop. In the strongest tested case, that subpart gives about $4.7$ times more loss reduction per stored byte than the whole block. Blocks 17 and 28 contain an attention subpart that lowers loss even though the whole-block change raises it. Smaller groups reveal these cases. They do not change the storage-efficient ordering. The best decoder subpart lowers loss by about $0.0003$ per million weights. A leading image-encoder block lowers it by about $0.0023$ per million weights. The measured plans therefore still choose image-encoder blocks first. Table~\ref{tab:app-granularity} shows eight representative blocks from the 13 tested.

\begin{table}[!ht]
  \centering
  \caption{\textbf{No decoder subpart is always the best precision target.} These eight representative blocks are a subset of the 13 tested. Entries are drops in faithfulness loss, and positive is better. The two subparts can move in opposite directions within one block.}
  \label{tab:app-granularity}
  \vspace{3pt}
  \setlength{\tabcolsep}{6pt}
  \begin{tabular}{lrrrrl}
    \toprule
    Block & Whole block & Token mixing & FFN & Parts minus whole & Larger gain \\
    \midrule
    dec.00 & .0087    & .0088    & $-.0033$ & $-.0032$ & attention   \\
    dec.01 & .0249    & $-.0001$ & .0275    & .0025    & FFN         \\
    dec.06 & .0197    & .0217    & $-.0039$ & $-.0020$ & attention   \\
    dec.08 & .0193    & .0100    & .0082    & $-.0011$ & shared      \\
    dec.10 & .0209    & .0213    & .0005    & .0009    & attention   \\
    dec.13 & .0223    & .0049    & .0122    & $-.0052$ & FFN         \\
    dec.17 & $-.0045$ & .0042    & $-.0046$ & .0041    & interaction \\
    dec.28 & $-.0119$ & .0016    & $-.0130$ & .0005    & FFN         \\
    \bottomrule
  \end{tabular}
\end{table}

\paragraph{The part ranking is not confirmed}
These tests use all 1,024 questions in the calibration pool. Their loss combines scaled PRS$_+$ with any increase in the negative log probability of the correct answer relative to 16-bit. This is not the filtered PRS in Table~\ref{tab:app-populations}. Near-zero image effects can dominate PRS$_+$. Rankings from two halves of the data correlate only $0.307$. The map describes the scoring rule in this run. It does not establish a stable causal order for every prompt. Such an order would require a filtered test on new calibration images.

\subsection{Does the Best Choice Vary by Question?}
\label{sec:app-routing}

\paragraph{Hindsight still appears useful after the tested link is destroyed}
The hindsight chooser selects the best precision plan after seeing each question's measured outcome. It cannot be deployed. A fixed plan uses one choice for every question. At 10\%, the hindsight advantage over the fixed plan is $1.3535$. Its 95\% interval is $[1.2142,1.5066]$ after resampling whole images. The recorded randomization test against zero gives $p<10^{-4}$. We then destroy the question-to-group link in two ways. Shuffling question outcomes within each group gives an advantage of $1.6616$. Shuffling group identities within each question gives $1.2309$. These controls keep the score distributions but remove the tested relationship. The real advantage must exceed both controls at every budget. It does not (Table~\ref{tab:app-null-gate}).

\begin{table}[!ht]
  \centering
  \caption{\textbf{Per-question hindsight gains stay below at least one shuffled control.} Each budget is the extra weight storage allowed for one question. Shuffle A preserves each group's score distribution. Shuffle B preserves each question's score distribution. Only one value is available for each shuffle and budget. Shuffle counts and seeds are unavailable.}
  \label{tab:app-null-gate}
  \vspace{3pt}
  \setlength{\tabcolsep}{6pt}
  \begin{tabular}{crrrrrr}
    \toprule
    \shortstack{Extra storage\\(\% of W4)} & \shortstack{One fixed\\plan} & \shortstack{Hindsight\\best} & \shortstack{Real\\advantage} & \shortstack{Shuffle A} & \shortstack{Shuffle B} & \shortstack{Both?} \\
    \midrule
    5\%  & .2595 & 1.3561 & 1.0966 & 1.4557 &  .9579 & no \\
    10\% & .3067 & 1.6602 & 1.3535 & 1.6616 & 1.2309 & no \\
    15\% & .3276 & 1.8734 & 1.5458 & 1.8313 & 1.4778 & no \\
    \bottomrule
  \end{tabular}
\end{table}

\paragraph{Most variation cannot be separated from noise}
Average differences among groups explain $0.1496\%$ of recovery variance. Average differences among questions explain $6.3301\%$. The combined question-by-group term and measurement noise explain $93.5203\%$. The top-five group sets share little. Their shared-over-union fraction averages $0.131$ and has median $0.111$. The globally best group is also best for only $0.879\%$ of questions. Entropy measures how widely the winning groups are spread. It is $4.12$ bits out of a possible $5.83$. Kendall rank agreement measures consistency across category rankings. It is $\tau=0.004$, where zero means no consistent ordering. These results show high diversity. The shuffled controls show that diversity alone does not establish repeatable question-specific structure.

\paragraph{All three selectors make the same decision}
A precision selector uses inexpensive W4 features to choose a precision profile before the answer is known. Two selectors learn this choice from data. They are a regularized linear model called ridge and a two-layer neural network. A third method uses a fixed Jensen--Shannon distance. It compares probability distributions and is not learned. Each selector adds a held-out error margin to its loss prediction. If no profile's upper prediction falls below the limit, the method assigns K3\_P2 and applies VCD. All three methods do this for every held-out calibration example. Across four profiles, their mean absolute prediction errors span $0.429$--$0.642$. A shown in Table~\ref{tab:app-router}, ``coverage'' is the share served without VCD. Coverage is $0\%$ at the W8-based loss limit $\epsilon_f=0.3240$. Before VCD, the mean faithfulness loss is $0.4075$. The fixed K3\_P3 profile has loss $0.3557$ on the same examples. The paired difference is $0.0518$ with 95\% CI $[-0.0065,0.1129]$ ($p=0.0816$). Randomly reassigning profiles with the same frequencies changes the result by exactly zero. Uniform W4 and W8 have losses $0.5502$ and $0.0699$.

\begin{table}[!ht]
  \centering
  \caption{\textbf{All three selectors assign K3\_P2, but VCD is applied to every example.} Results use held-out calibration examples. Extra bytes are the mean used for one question. Storing all candidates costs 19.6\% more than W4. Every upper prediction exceeds the W8-based limit.}
  \label{tab:app-router}
  \vspace{3pt}
  \small
  \setlength{\tabcolsep}{4pt}
  \begin{tabular}{lrrrrl}
    \toprule
    \shortstack[l]{\\How weight bits\\are chosen} & \shortstack[r]{Mean extra\\bytes used\\(\% of W4)} & \shortstack[r]{\\Loss (lower\\is better)} & \shortstack[r]{\\\\Coverage (\%)} & \shortstack[r]{\\Gap to\\hindsight} & \shortstack[l]{\\\\Choice} \\
    \midrule
    W4                         &   0.0 & .5502          & --  & --    & fixed            \\
    Best fixed K3\_P3          &   7.0 & \textbf{.3557} & --  & --    & fixed            \\
    Random reassignment        &   9.4 & .4075          & --  & --    & same frequencies \\
    Distribution-distance rule &   9.4 & .4075          & 0.0 & .2229 & 100\% K3\_P2     \\
    Linear ridge model         &   9.4 & .4075          & 0.0 & .2229 & 100\% K3\_P2     \\
    Two-layer neural net       &   9.4 & .4075          & 0.0 & .2229 & 100\% K3\_P2     \\
    Hindsight best plan        &   9.4 & .1847          & --  & .0000 & cannot deploy    \\
    W8                         & 100.0 & .0699          & --  & --    & fixed            \\
    \bottomrule
  \end{tabular}
\end{table}

\paragraph{A looser loss limit includes more questions by changing the target}
As $\epsilon_f$ rises, more questions have actual loss below it even with perfect hindsight. This gives the highest possible share below the limit before prediction error. The W8-based limit is strict. The two learned selectors and the fixed distance rule all end with loss near $0.408$. Their profile-wise mean absolute errors lie between $0.43$ and $0.64$. Raising the limit can include more questions. It does not show that W4 features are informative. It also no longer meets the original W8 standard.

\begin{figure}[!ht]
\centering
\begin{minipage}[t]{0.44\linewidth}
\vspace{0pt}
\centering
\footnotesize\textbf{(a) More questions meet a looser loss limit}\\[-2pt]
\includegraphics[width=\linewidth]{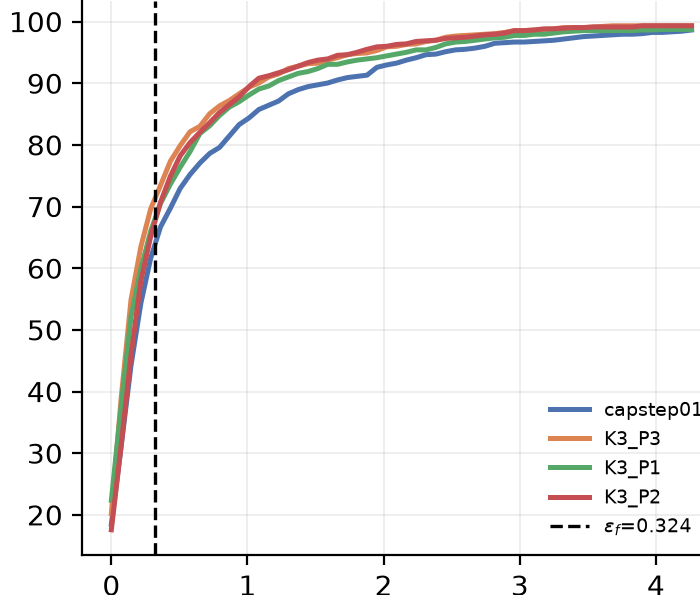}
\end{minipage}\hfill
\begin{minipage}[t]{0.52\linewidth}
\vspace{0pt}
\centering
\includegraphics[width=\linewidth]{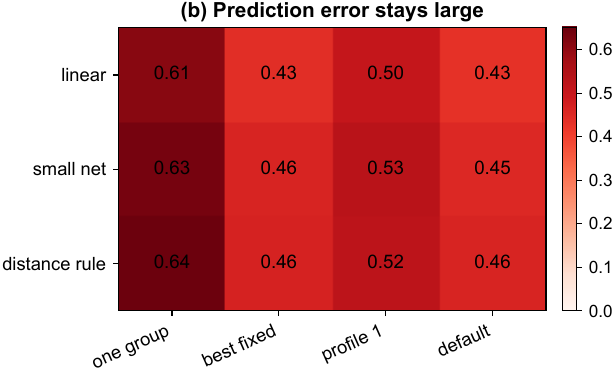}
\end{minipage}
\caption{\textbf{A looser W8 loss limit includes more questions but does not improve prediction.} Left: even hindsight needs a looser limit. Right: errors of $0.43$--$0.64$ are about as large as the W4-to-W8 loss span, $0.55$ to $0.07$. In the left plot, ``risk'' means faithfulness loss.}
\label{fig:app-risk-threshold}
\end{figure}
\FloatBarrier

\subsection{Fixed Precision Plans and General-Skill Evaluation}
\label{sec:app-profiles}

\paragraph{The small profile menu yields only fixed candidate plans}
K-medoids is a clustering method that chooses representative plans. It groups the hindsight choices into three mixed W4/W8 profiles (Table~\ref{tab:app-profiles}). K3\_P1 through K3\_P3 are experiment identifiers. K3\_P3 has the lowest calibration-set faithfulness loss, $0.4003$. It uses 23 groups and 7.0\% extra storage. K3\_P2 uses more extra storage but has higher loss $0.4297$. More precision is therefore not always better. The tested selectors assign K3\_P2 when no profile's upper prediction falls below the limit. Hindsight choices are inflated by noise, and every selector assigns K3\_P2. These profiles are fixed candidates. They do not show that different questions need different profiles.

\begin{table}[!ht]
  \centering
  \caption{\textbf{Representative fixed plans reduce loss on the calibration evaluation used to build the profiles.} ``Skill change'' is the percentage-point change on general capability evaluations. Lower loss is better.}
  \label{tab:app-profiles}
  \vspace{3pt}
  \small
  \setlength{\tabcolsep}{4pt}
  \begin{tabular}{lrrrrrr}
    \toprule
    Precision plan & W8 groups & \shortstack{Extra storage\\(\% of W4)} & Loss & PRS$_+$ & Calibration accuracy & Skill change \\
    \midrule
    W4             &  0 & 0.0 & .6021          & .2599 & 81.2 & .05     \\
    P0 (capstep01) &  1 &  .2 & .5694          & .2365 & 81.4 & $-.02$  \\
    K3\_P1         & 14 & 8.3 & .4629          & .1830 & 81.4 & .34     \\
    K3\_P2         & 20 & 9.4 & .4297          & .1865 & 81.2 & .21     \\
    K3\_P3         & 23 & 7.0 & \textbf{.4003} & .1679 & 80.9 & $-.03$  \\
    \bottomrule
  \end{tabular}
\end{table}
\FloatBarrier

\paragraph{Larger budgets continue to change the plan}
We allow 5\%, 10\%, or 15\% more weight storage than W4 for one question. Selection by average-token score chooses 21, 23, and 24 groups. The corresponding losses are $0.4165$, $0.3532$, and $0.3257$. Each budget produces a different plan. Selection by general-skill score is better at 5\% and 10\%, with losses $0.3732$ and $0.3281$ as shown in Table~\ref{tab:app-budget}. The two losses are nearly equal at 15\%, where the general-skill value is $0.3266$. The results do not show saturation at 5\%. They also do not show that average-token-score selection is best. The 10\% average-token plan contains 21 vision blocks and decoder blocks 1 and 13. The 15\% plan also adds decoder block 10. 

\begin{table}[!ht]
  \centering
  \caption{\textbf{Larger per-question storage budgets continue to change the plan.} These losses come from the calibration-set budget sweep. They cannot be compared directly with the full-POPE losses in Table~\ref{tab:app-pope-grid}.}
  \label{tab:app-budget}
  \vspace{3pt}
  \begin{tabular}{clrrr}
    \toprule
    \shortstack{Storage budget\\per question} & Selection target & W8 groups & \shortstack{Calculated extra storage\\(\% of W4)} & Loss \\
    \midrule
    5\%   & average-token score & 21 &  3.9 & .4165          \\
          & general-skill score & 22 &  4.2 & \textbf{.3732} \\
    10\%  & average-token score & 23 &  9.8 & .3532          \\
          & general-skill score & 23 &  9.8 & \textbf{.3281} \\
    15\%  & average-token score & 24 & 12.8 & \textbf{.3257} \\
          & general-skill score & 25 & 13.1 & .3266          \\
    \bottomrule
  \end{tabular}
\end{table}

\paragraph{W4 already meets the general-skill criterion}
We test MME and MMBench as multimodal tasks. We test ARC and HellaSwag with a fixed neutral image. W4 changes average accuracy by $+0.05$ percentage points. No task has a statistically resolved violation. The three task changes are $+0.3448$, $+0.4000$, and $-0.6000$ points as shown in Table~\ref{tab:app-capability}. The worst raw category falls by $-3.33$ points. Task categories contain 40--250 examples. Those with negative changes contain 60--125 examples, and the drops are within two task-level standard errors. The smallest valid protection set is therefore empty. The recorded capstep01 plan changes vision block 13 to W8 only because the procedure required one candidate. The general-skill criterion does not require this group.

\begin{table}[!ht]
  \centering
  \caption{\textbf{Every candidate meets the skill criterion, including the empty protection set.} Skill score is the Jensen--Shannon distance from the 16-bit answer distribution. Lower is better. Accuracy changes are percentage points from the 16-bit model. A resolved drop must exceed both 2 points and two standard errors for that task's sample size.}
  \label{tab:app-capability}
  \vspace{3pt}
  \setlength{\tabcolsep}{4.5pt}
  \begin{tabular}{lrrrrrr}
    \toprule
    Protected groups & Count & \shortstack{Extra storage\\(\% of W4)} & Skill score $\downarrow$ & \shortstack{Avg. accuracy\\change} & \shortstack{Worst raw\\change} & \shortstack{Resolved\\drops} \\
    \midrule
    W4, empty set &  0 & 0.0 & .0088 &  .05   & $-3.33$ & 0 \\
    Top 1         &  1 &  .2 & .0087 & $-.02$ & $-3.33$ & 0 \\
    Top 2         &  2 &  .4 & .0086 & $-.15$ & $-3.33$ & 0 \\
    Top 3         &  3 &  .6 & .0086 &  .04   & $-3.20$ & 0 \\
    Top 4         &  4 &  .7 & .0084 &  .03   & $-3.20$ & 0 \\
    Top 6         &  6 & 1.1 & .0081 &  .16   & $-3.20$ & 0 \\
    Top 8         &  8 & 1.5 & .0080 &  .22   & $-3.20$ & 0 \\
    Top 10        & 10 & 1.9 & .0077 &  .30   & $-3.20$ & 0 \\
    Top 12        & 12 & 2.2 & .0076 &  .31   & $-3.20$ & 0 \\
    \bottomrule
  \end{tabular}
\end{table}

\FloatBarrier

\paragraph{A fixed mixed-precision plan lowers the measured loss}
On held-out calibration examples, capstep01 lowers faithfulness loss by $0.0318$ from W4. Moving to the K3\_P3 profile lowers it by another $0.1627$. Replacing K3\_P3 with the fixed 10\% average-token plan changes loss by only $-0.0025$. Per-question hindsight reaches $0.1847$. Section~\ref{sec:app-routing} shows that this value is inflated by choosing the best noisy outcome. The measured gain comes from one fixed mixed-precision plan. The result does not establish a necessary skill constraint or a deployable adaptive bound.

\subsection{Do Any Answers or Captions Improve?}
\label{sec:app-behavior}

\paragraph{POPE accuracy differences are unresolved for all 12 precision plans}
POPE asks balanced yes/no questions about whether an object appears in an image. F1 balances precision and recall for yes answers. Across Table~\ref{tab:app-pope-grid}, adversarial F1 spans $82.5$--$83.2$. Overall accuracy spans $85.6$--$86.1$.
The fixed 10\% average-token plan lowers all-question PRS$_+$ from W4 by $0.0650$. Its 95\% CI is $[-0.0868,-0.0424]$. Adversarial accuracy changes by $-0.10$ points, with CI $[-0.70,+0.50]$. A paired McNemar correctness test gives $p=0.7493$. Compared with putting the vision encoder at W8, the plan's PRS$_+$ difference is $+0.0085$. Its CI is $[-0.0079,0.0264]$. Compared with the first eight decoder blocks at W8, it improves PRS$_+$ by $-0.0247$, with CI $[-0.0476,-0.0008]$. Per-question selection differs from the fixed 10\% average-token plan by $+0.0046$.

\begin{table}[!ht]
  \centering
  \caption{\textbf{Internal answer scores move while final POPE answers remain nearly constant.} PRS$_+$ and loss use all 9,000 questions. Average weight bits are calculated from parameter counts. The 16-bit model's value is 16.}
  \label{tab:app-pope-grid}
  \vspace{3pt}
  \footnotesize
  \setlength{\tabcolsep}{3pt}
  \begin{tabular}{lrrrrrrrr}
    \toprule
    Precision plan & \shortstack{F1\\random} & \shortstack{F1\\popular} & \shortstack{F1\\adversarial} & Accuracy & Yes (\%) & PRS$_+$ & Loss & \shortstack{Avg. weight\\bits} \\
    \midrule
    16-bit reference               & 86.1 & 85.0 & 83.0 & 86.0 & 41.7 & .000 & .0000 & 16.00 \\
    Uniform W8                     & 86.2 & 85.2 & 83.2 & 86.1 & 41.7 & .032 & .0761 &  8.00 \\
    Uniform W4                     & 86.5 & 84.9 & 82.9 & 85.9 & 42.1 & .241 & .5141 &  4.00 \\
    Vision encoder W8              & 86.1 & 84.6 & 82.6 & 85.6 & 42.1 & .167 & .2952 &  4.18 \\
    Vision encoder + connector W8  & 86.1 & 84.6 & 82.7 & 85.7 & 42.1 & .163 & .2846 &  4.19 \\
    First 8 decoder blocks W8      & 86.6 & 84.9 & 82.8 & 85.9 & 42.3 & .201 & .4577 &  4.95 \\
    Connector W8                   & 86.2 & 84.6 & 82.5 & 85.7 & 42.0 & .237 & .5072 &  4.01 \\
    Fixed 5\% average-token plan   & 86.1 & 84.4 & 82.7 & 85.6 & 42.0 & .191 & .3347 &  4.16 \\
    Fixed 10\% average-token plan  & 86.4 & 84.6 & 82.8 & 85.8 & 42.4 & .176 & .3073 &  4.39 \\
    Fixed 15\% average-token plan  & 86.4 & 84.6 & 82.8 & 85.8 & 42.3 & .164 & .2891 &  4.51 \\
    capstep01 only                 & 86.5 & 84.8 & 82.7 & 85.8 & 42.3 & .242 & .4998 &  4.01 \\
    Per-question chooser           & 86.1 & 84.9 & 82.8 & 85.8 & 42.0 & .181 & .3282 &  4.38 \\
    \bottomrule
  \end{tabular}
\end{table}

\FloatBarrier

\paragraph{Open-ended captions also show no resolved change}
CHAIR measures unsupported object mentions in generated captions. An object counts as present if it appears in either the COCO instance labels or any of the five reference captions. We test 500 COCO captions. The sentence-level error rate CHAIR$_s$ is $24.0\%$ for the 16-bit model and $24.6\%$ for W4. Their paired difference is $-0.60$ points with 95\% CI $[-4.00,2.80]$ ($p=0.7728$). K3\_P3 has the lowest rate at $23.2\%$. Its difference from W4 is $-1.40$ points, with CI $[-4.60,1.60]$ ($p=0.4154$). As shown in Table~\ref{tab:app-chair}, object recall stays between $61.9\%$ and $63.9\%$. Caption length stays between $49.3$ and $49.6$ words. Objects per caption stay between $4.98$ and $5.10$. The unresolved result is not explained by much shorter captions. Each image has one caption. Greedy decoding chooses the highest-scoring next token at each step. We use a hand-built list of COCO object synonyms and not the original CHAIR vocabulary. The absolute error rates therefore cannot be compared directly with published CHAIR numbers.

\begin{table}[!ht]
  \centering
  \caption{\textbf{Differences in unsupported object mentions are unresolved.} Error rates and recall are percentages. Changes and intervals are percentage points. Intervals come from 10,000 paired image-level resamples.}
  \label{tab:app-chair}
  \vspace{3pt}
  \setlength{\tabcolsep}{5pt}
  \begin{tabular}{lrrrrrl}
    \toprule
    Precision plan & Sent. err. & Wrong obj. & Recall & Words & Obj./cap. & $\Delta$ sent. err. [95\% CI] \\
    \midrule
    16-bit reference & 24.0 & 6.9 & 63.5 & 49.4 & 5.04 & $-.60$ [$-4.00,2.80$]  \\
    W8               & 24.6 & 7.0 & 63.9 & 49.4 & 5.08 & $.00$ [$-3.40,3.40$]   \\
    W4               & 24.6 & 7.3 & 61.9 & 49.6 & 4.98 & reference              \\
    K3\_P1           & 25.0 & 7.5 & 63.2 & 49.3 & 5.08 & $.40$ [$-2.40,3.40$]   \\
    K3\_P2           & 25.2 & 7.6 & 63.1 & 49.4 & 5.10 & $.60$ [$-2.60,3.80$]   \\
    K3\_P3           & 23.2 & 6.5 & 63.1 & 49.4 & 5.08 & $-1.40$ [$-4.60,1.60$] \\
    capstep01        & 24.6 & 7.3 & 62.2 & 49.6 & 5.00 & $.00$ [$-1.60,1.60$]   \\
    \bottomrule
  \end{tabular}
\end{table}

\paragraph{A second model repeats the ordering with a smaller shift}
Qwen2.5-VL-7B-Instruct uses its own cutoff, $\tau_d=0.45299$. Its W4 PRS is $0.01397$, compared with $0.00017$ for W8, as shown in Table~\ref{tab:app-qwen}. Combined F1 is $88.62$ for W4 and $88.94$ for W8. The higher PRS at 4 bits matches LLaVA. LLaVA's filtered W4 mean is larger at $0.07596$. The visual-response ratio (VRR) compares the size of the lower-precision image response with the 16-bit response. A value of one means equal size. Qwen's W4 VRR is $0.9985$. The directional reversion rate (DRR) is the share of changed answers that match the image-replacement answer. Qwen's W4 DRR is $0.5766$, compared with $0.5372$ after shuffling ($p=0.1979$). Qwen provides a secondary diagnostic evaluation. Images are fixed at $448\!\times\!448$ with 256 visual tokens. LLaVA uses 576 visual tokens. We do not test mixed-precision plans on Qwen. The experiment cannot separate model-design effects from token-count effects.

\begin{table}[!ht]
  \centering
  \caption{\textbf{Both models show W4 above W8, but Qwen's shift is smaller.} Mean PRS uses each model's filtered questions. F1, image-response ratio, and flip rates use all questions. LLaVA uses cutoff $\tau_d=0.302$. Qwen uses $0.45299$. Their score sizes are therefore not directly comparable.}
  \label{tab:app-qwen}
  \vspace{3pt}
  \setlength{\tabcolsep}{4.5pt}
  \begin{tabular}{llrrrrrr}
    \toprule
    Model & Weight bits & F1 & Avg. PRS & \shortstack{Image-response\\ratio} & \shortstack{Matching\\flips} & \shortstack{Shuffled\\flips} & $p$ \\
    \midrule
    LLaVA-1.5-7B  & W8 & 84.81 & .00940 & 1.0078 & .4545 & .4563 & .9740 \\
                  & W4 & 84.72 & .07596 & 1.0578 & .4264 & .4269 & .7225 \\
    Qwen2.5-VL-7B & W8 & 88.94 & .00017 & 1.0000 & .2647 & .4451 & .9980 \\
                  & W4 & 88.62 & .01397 &  .9985 & .5766 & .5372 & .1979 \\
    \bottomrule
  \end{tabular}
\end{table}
\FloatBarrier

\subsection{Contrastive Decoding, Examples, and Storage Estimates}
\label{sec:app-vcd}

\paragraph{A two-image correction improves one score and hurts another}
Visual contrastive decoding (VCD) runs the model on two images. One is the real image $I$. The other is a noise-corrupted copy $I'$. VCD combines their answer scores as $(1+\alpha)z(I)-\alpha z(I')$. The number of diffusion steps controls the amount of corruption. The value $\alpha$ controls correction strength. The cutoff $\beta=0.1$ removes answers that the real-image run considers too unlikely. With 999 steps, increasing $\alpha$ from 0 to 1 moves all-question PRS from $0.0458$ to $-0.6884$. Accuracy falls from $85.92\%$ to $83.19\%$. The yes-rate rises from $42.1\%$ to $57.8\%$. As shown in Table~\ref{tab:app-vcd}, F1 changes by only $-0.32$ points, from $84.72$ to $84.41$. The $2.73$-point drop applies to accuracy. F1 peaks at $\alpha=0.25$, where it is $85.74$. PRS continues to fall through $\alpha=1$. At the same time, PRS$_+$ rises from $0.241$ to $1.243$ because the score distribution becomes wider. Mean PRS and PRS$_+$ therefore favor different settings.

\begin{table}[!ht]
  \centering
  \caption{\textbf{PRS and F1 favor different correction strengths.} Every row uses $\beta=0.1$. Missing statistics from the corruption sweep are shown as dashes. They are not estimated. PRS uses all questions.}
  \label{tab:app-vcd}
  \vspace{3pt}
  \setlength{\tabcolsep}{4.2pt}
  \begin{tabular}{lrrrrrr}
    \toprule
    Image correction & Accuracy & F1 & Yes (\%) & Avg. PRS & PRS$_+$ & Loss \\
    \midrule
    W4, no VCD              & 85.92 & 84.72          & 42.1 &  .0458   & .241  &  .514 \\
    999 steps, $\alpha=.25$ & --    & \textbf{85.74} & 48.3 & $-.138$  & --    &  .987 \\
    999 steps, $\alpha=.50$ & --    & 84.98          & 52.1 & $-.321$  & --    & 1.889 \\
    999 steps, $\alpha=1$   & 83.19 & 84.41          & 57.8 & $-.688$  & 1.243 & 4.122 \\
    500 steps, $\alpha=.25$ & --    & 84.9           & 45.4 & $-.140$  & --    &  .769 \\
    500 steps, $\alpha=.50$ & --    & 85.1           & 48.4 & $-.327$  & --    & 1.333 \\
    500 steps, $\alpha=1$   & --    & 84.6           & 51.0 & $-.699$  & --    & 2.712 \\
    \bottomrule
  \end{tabular}
\end{table}

\paragraph{The corrupted-image run explains the strong shift toward yes}
With 999 corruption steps, the corrupted-image run alone answers yes on $0.3\%$ of questions. Its F1 is $0.8$. Subtracting its scores pushes the combined answer strongly toward yes. At 500 steps, the corrupted-image run answers yes on $11.4\%$ and has F1 $29.8$. Average PRS at matched $\alpha$ changes little. The proposed selective policy uses VCD on $100\%$ of questions because no calibrated upper prediction falls below the measured-loss limit. Its outputs exactly match always-on VCD. This experiment therefore tests an always-on two-image correction. It does not demonstrate selective computation.

\begin{table}[!ht]
  \centering
  \caption{\textbf{Selective and always-on VCD are identical because no upper prediction falls below the loss limit.} Yes, Positive, and VCD used are percentages. Every VCD row uses $\alpha=1$ and $\beta=0.1$. It also uses the same weight precision for the real and corrupted images.}
  \label{tab:app-vcd-grid}
  \vspace{3pt}
  \small
  \setlength{\tabcolsep}{3.2pt}
  \begin{tabular}{lrrrrrrrr}
    \toprule
    Precision and correction plan & Acc. & F1 & Yes & Avg. PRS & Positive & PRS$_+$ & Loss & VCD used \\
    \midrule
    16-bit reference        & 86.0 & 84.7 & 41.7 &  .000   &  0 & .000  &  .000 &   0 \\
    W4, no VCD              & 85.9 & 84.7 & 42.1 &  .046   & 71 & .241  &  .514 &   0 \\
    W8, no VCD              & 86.1 & 84.8 & 41.7 &  .001   & 62 & .032  &  .076 &   0 \\
    W4 + always-on VCD      & 83.2 & 84.4 & 57.8 & $-.688$ & 32 & 1.243 & 4.122 & 100 \\
    Fixed K3\_P3            & 85.6 & 84.4 & 42.2 &  .044   & 66 & .161  &  .295 &   0 \\
    Fixed K3\_P3 + VCD      & 83.0 & 84.3 & 57.8 & $-.696$ & 30 & 1.236 & 4.100 & 100 \\
    Chooser, no VCD         & 85.8 & 84.6 & 42.0 &  .052   & 71 & .181  &  .328 &   0 \\
    Chooser + selective VCD & 83.2 & 84.3 & 57.2 & $-.695$ & 31 & 1.195 & 3.905 & 100 \\
    Chooser + always-on VCD & 83.2 & 84.3 & 57.2 & $-.695$ & 31 & 1.195 & 3.905 & 100 \\
    \bottomrule
  \end{tabular}
\end{table}

\paragraph{The clearest examples are rare and use one fixed higher-precision profile}
Only 66 of 9,000 POPE questions ($0.73\%$) follow the complete pattern, as shown in Table~\ref{tab:app-vcd-grid}. The 16-bit answer is correct. Replacing image tokens reverses it. W4 then matches the replacement answer. Each question type contributes 22 cases. The fixed K3\_P2 profile restores 45 of 66 cases ($68.2\%$). The selectors also assign K3\_P2 to every question, so the restoration cannot be attributed to routing. The selected examples contain three truck questions and three spoon questions. Four are restored. Two spoon cases remain wrong. The figure illustrates the pattern. It does not show prevalence or successful question-specific selection.

\FloatBarrier
\clearpage

\end{document}